\pdfoutput=1

\documentclass[11pt]{article}

\usepackage[final]{acl}

\usepackage{times}
\usepackage{latexsym}
\usepackage{amsmath}

\usepackage[T1]{fontenc}

\usepackage[utf8]{inputenc}

\usepackage{microtype}

\usepackage{inconsolata}
\usepackage{enumitem}
\usepackage{tabularx}
\usepackage{amssymb}
\usepackage{graphicx}
\usepackage{booktabs}
\usepackage{siunitx}  
\usepackage{multirow}

\usepackage{makecell} 

\usepackage{xspace} 

\newcommand{\gigabase}{GigaChat-A3B-base\xspace}
\newcommand{\gigainstruct}{GigaChat-A3B-instruct\xspace}
\newcommand{\gigainstructdpo}{GigaChat-A3B-instruct 1.5\xspace}

\newcommand{\gigapronew}{GigaChat 2 Pro\xspace}
\newcommand{\gigamaxnew}{GigaChat 2 MAX\xspace}
\newcommand{\giganew}{GigaChat 2\xspace}

\title{GigaChat Family: Efficient Russian Language Modeling Through Mixture of Experts Architecture}

\author{
    GigaChat team \\ SaluteDevices / Moscow \\
    \small{
        \textbf{Correspondence:} \href{mailto:minkin.fyodor@gmail.com}{minkin.fyodor@gmail.com}
    }
}

\begin{document}
\maketitle
\begin{abstract}
Generative large language models (LLMs) have become crucial for modern NLP research and applications across various languages. However, the development of foundational models specifically tailored to the Russian language has been limited, primarily due to the significant computational resources required. This paper introduces the GigaChat family of Russian LLMs, available in various sizes, including base models and instruction-tuned versions. We provide a detailed report on the model architecture, pre-training process, and experiments to guide design choices. In addition, we evaluate their performance on Russian and English benchmarks and compare GigaChat with multilingual analogs. The paper presents a system demonstration of the top-performing models accessible via an API, a Telegram bot, and a Web interface. Furthermore, we have released three open GigaChat models in open-source~\footnote{\url{https://huggingface.co/ai-sage}}, aiming to expand NLP research opportunities and support the development of industrial solutions for the Russian language.
\end{abstract}

\section{Introduction}
\label{sec:intro}
The rapid advancement of generative large language models (LLMs) has significantly transformed the landscape of natural language processing (NLP), enabling innovative research and applications across multiple languages. However, developing foundation and post-trained models for the Russian language is still a significant challenge. This resource-intensive task hinders progress in the field and fails to address the cultural specifics of the Russian language and culture.

In response to this gap, we introduce the GigaChat family of Russian LLMs, created from scratch, which encompasses a variety of sizes, including both pre-trained and instruction-tuned versions. This paper describes our experience creating a model family based on the mixture of experts (MoE) architecture, the experiments in training such an architecture, and the description of the new tokenizer designed for the Russian language. Furthermore, we thoroughly evaluate the model's performance on Russian and English benchmarks and tests. This paper not only highlights the strengths of GigaChat in comparison to existing multilingual models but also offers a practical demonstration of our top-performing proprietary models through accessible interfaces such as an API, a Telegram bot, and a web application. By releasing three open versions of the GigaChat models as open-source resources, we aim to encourage further research in natural language processing (NLP) and support the ongoing development of industrial applications tailored to the Russian language.

Our contributions are as follows:
\begin{itemize}
\item We introduce the first family of foundation and post-trained models specifically designed for the Russian language, based on the Mixture of Experts (MoE) architecture. Three of these models are available in open-source (including their variations in int8 and bf16 formats)~\footnote{Under the MIT license, commercial/non-commercial use, re-hosting, and fine-tuning are permitted without restrictions.}.

\item We present experimental results and metrics on various benchmarks, demonstrating that our models are comparable to the state-of-the-art (SOTA) models of similar sizes among existing open-source models.

\item We also share our experiments with the MoE concentration mechanism and provide code for MoE expert control.

\item We release the Telegram bot and the System demo Web interface~\footnote{The \href{https://youtu.be/TDD9av314XY}{video demonstration} is available on YouTube.} for our most advanced model.
\end{itemize}

\section{Related Work}
\label{sec:reelated_work}

\begin{figure}[t]
  \includegraphics[width=\columnwidth]{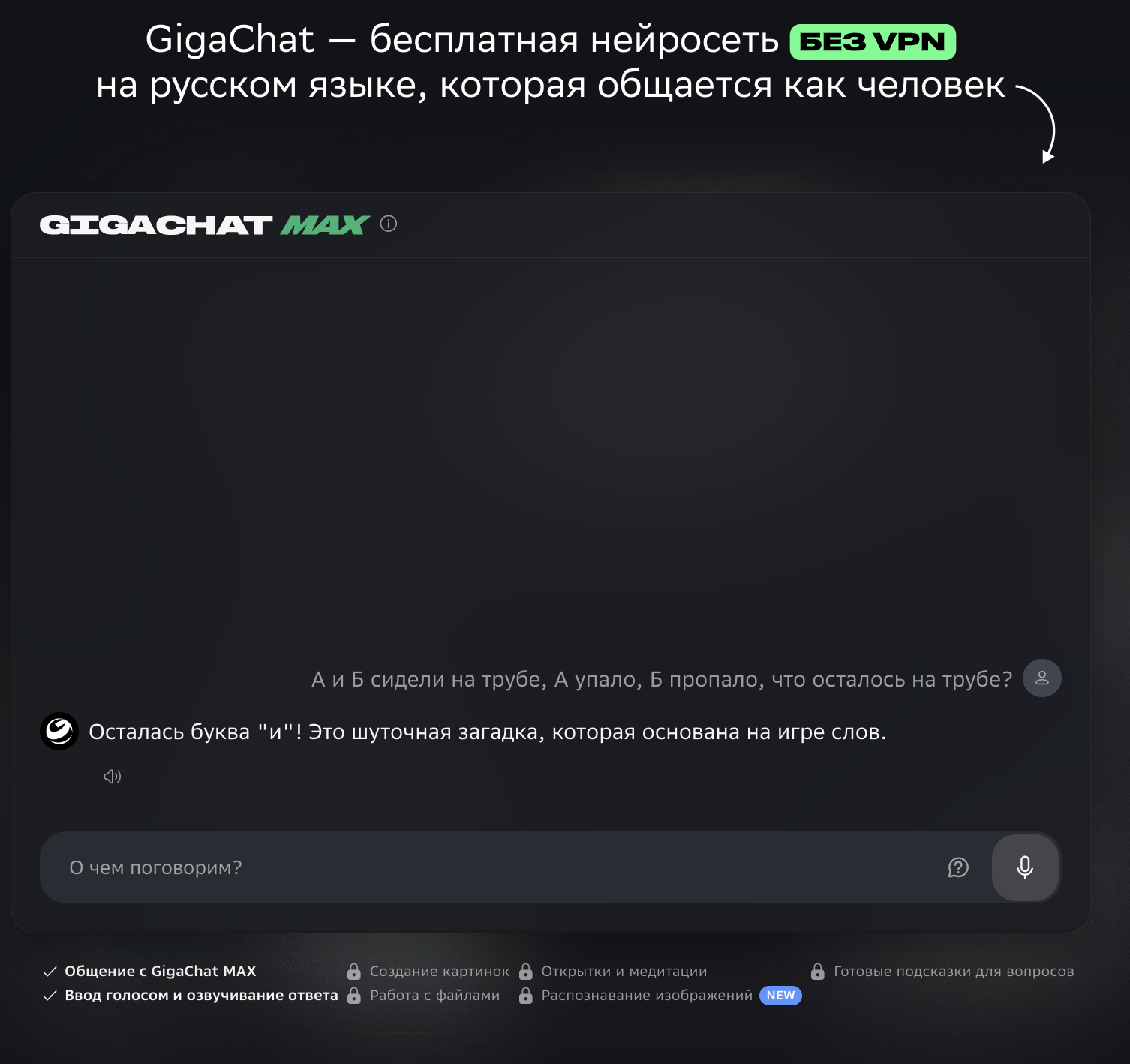}
  \caption{A screenshot of the system demo for the open  Web demo of the GigaChat Max. To access more features of GigaChat, registration is required.}
  \label{fig:demo_screen}
\end{figure}

\noindent\textbf{MoE architecture}
\noindent Sparse MoE models have gained significant attention in recent years~\cite{cai2024surveymixtureexperts} due to their capacity for efficient scaling while maintaining computational effectiveness. The foundational work \citet{DBLP:conf/iclr/ShazeerMMDLHD17} introduced the sparse MoE layer, demonstrating its effectiveness in training large-scale language models in application to LSTM-based architectures. More recently, Mixtral~\cite{Jiang2024MixtralOE} set a new SOTA for MoE-based LLMs with 47 billion total parameters but only 13 billion active parameters, outperforming dense models such as LLaMA 2 70B. Another notable contribution, DeepSeek MoE~\cite{Dai2024DeepSeekMoETU}, explored modifications to MoE architecture by increasing the number of experts while reducing their sizes and adding shared experts that are always activated, improving expert specialization and overall model performance.

\noindent\textbf{Russian generative LLMs}. Pre-trained open models for the Russian language remain scarce. The work of \citet{zmitrovich2024family} introduces a collection of 13 Russian Transformer-based language models, which include encoder architectures (ruBERT, ruRoBERTa, ruELECTRA), decoder architectures (ruGPT-3), and encoder-decoder architectures (ruT5, FRED-T5). However, even the latest generative models, such as ruGPT-3.5~\footnote{\url{https://mera.a-ai.ru/ru/submits/11273}}, demonstrate subpar performance on benchmarks like the MERA SOTA instruction models~\cite{fenogenova2024mera}. Most SOTA models, mainly those available as open-source, are either English-based or multilingual (e.g., Qwen, Mistral, and their Russian-adapted variants~\footnote{\href{https://huggingface.co/t-tech/T-pro-it-1.0}{T-pro-it-1.0}, \href{https://huggingface.co/msu-rcc-lair/RuadaptQwen2.5-32B-Instruct}{RuadaptQwen2.5-32B-Instruct}, \href{https://huggingface.co/ZeroAgency/Zero-Mistral-Small-24B-Instruct-2501}{Zero-Mistral-Small-24B}}), which have been post-trained on Russian texts. Among the Russian proprietary models, only a few exist, such as Cotype by MTS AI and the YandexGPT family~\footnote{\url{https://ya.ru/ai/gpt-4}}, both of which lack transparency regarding their training methodologies and architectural details and are not fully pre-trained on Russian texts. To bridge this gap and address the need for high-performing, Russian-focused generative models that rival their multilingual counterparts, we introduce the \textbf{GigaChat} family.

\section{GigaChat Family}
\label{sec:giga}

\subsection{Overview}
\label{subsec:overview}
The GigaChat family is the first collection of foundation and post-trained models specifically designed and pre-trained \underline{from scratch} for the Russian language. The initial version~\footnote{It is noteworthy that the three open models were previously also available through an API, and they continue to receive regular enhancements and improvements.} of the GigaChat family employs the MoE architecture that we now release in open-source: base model, instructed version, and aligned with Direct Preference Optimization (DPO)~\cite{rafailov2023direct}. Advanced proprietary models — Lite, Pro, and MAX — are continually updated and accessible through a user API and a dedicated Telegram bot, ensuring ongoing improvements and enhanced usability.

\subsection{System demo}
\label{subsec:system_demo}
The GigaChat models support a versatile user interaction system, offering free access through a Telegram bot and a Web demo interface~\footnote{\url{https://giga.chat/}}. The Web version contains the advanced proprietary model, GigaChat Max~\footnote{The API for the system demo is updating to the latest versions; we are reporting the version of GigaChat 2 as of March 2025.} Max allows users to engage in conversations by submitting text prompts in both Russian and English, all within a predefined character limit. The screenshot in Figure~\ref{fig:demo_screen} illustrates the interface of the free version, which offers two primary features: 1) chatting capability and 2) audio ASR input via GigaAM~\footnote{\url{https://github.com/salute-developers/GigaAM}}. The full version of the interface is available only after registration and includes additional functionalities such as file processing and predefined prompts for various use cases.

The key features of the Telegram bot (@gigachat\_bot) include an interactive chatbot that engages users in conversation and the capability to invoke the Kandinsky model~\cite{arkhipkin2024kandinsky} for image generation based on user prompts. Additionally, the bot offers a variety of predefined user prompts and can process files.

\subsection{Open models}
\label{subsec:open_models}
In this section, we explain the choice of the architecture and all the parts of the models creation, starting with the pre-trained base model.
\begin{table*}[ht!]
\scriptsize
\centering
\setlength{\tabcolsep}{3pt}
\begin{tabular}{lccccccccc}
\toprule
\textbf{Model} & \textbf{Architecture} & \textbf{Parameters} & \textbf{Hidden Layers} & \textbf{Shared experts} & \textbf{Routed experts} & \textbf{KV Heads} & \textbf{Heads} & \textbf{Context Length} \\
\midrule
\gigabase & MoE & 20B & 28 & 2 & 64 & 8 & 16 & 131k \\
\bottomrule
\end{tabular}
\caption{Summary of the \gigabase model architecture configurations.}
\label{tab:architecture}
\end{table*}


\subsubsection{Models architecture}
\label{subsubsec:architecture}

The \gigabase model leverages a MoE architecture with 20 billion total parameters, of which approximately 3.3 billion are activated per forward pass (see  Table~\ref{tab:architecture}). In our experiments using the same data, the MoE design demonstrates significant efficiency gains, including double the training speed and a 40\% reduction in inference latency compared to similarly sized dense models, such as 8B LLaMA 3.

The efficiency stems from block-sparse computation using optimized STK Triton kernels rather than Megablocks and selective activation checkpointing, reducing computational requirements by 40\% versus a 7B dense model while processing 1 trillion tokens. These optimizations eliminate the need for expert parallelism while maintaining model performance. 
The architecture replaces standard MLP blocks with MoE layers (except the first layer, which uses a gated MLP due to token distribution challenges). Each MoE block employs multiple experts and an unnormalized router to promote specialization, following insights from DeepSeek MoE. The intermediate dimension is expanded to 14,336 (as in Mistral 7B~\cite{jiang2023mistral7b}) to enhance capacity, and experts are shared across layers to improve parameter efficiency. This combination of sparse computation, expert sharing, and optimized routing enables high throughput with reduced resource consumption, making the model scalable for large-scale training and inference.

Section~\ref{subsec:details} of the Appendix describes the training process details.

\subsubsection{Pre-train}

The base model was trained using a constant multi-step learning rate scheduler with warmup. The scheduler included a warmup period of 2000 batches, after which four learning rate decay steps took place at 30\%, 60\%, 90\%, and 98\% of the total training duration. At these milestones, the learning rate was reduced by multiplying by factors of 0.25, 0.0625, 0.015625, and 0.00390625 (i.e., $(0.25)^1$, $(0.25)^2$, $(0.25)^3$, and $(0.25)^4$, respectively). The initial learning rate was set to 1e-4. The training process used a global batch size of approximately 16 million tokens (2048 sequences with 8192 tokens per sequence) and accumulated 9.5 trillion tokens across 8k pre-training steps.

After the initial training step, we conducted a context extension in two stages: first to 32K and then to 128K. To improve performance with the extended context, we adjusted the base for RoPE embeddings~\cite{su2024roformer} using the ABF approach~\cite{xiong2023effectivelongcontextscalingfoundation}. For each training stage, we utilized the following values: 10K for the initial 8K context, 300K for 32K, and 1.4M for 128K. The model employed a constant learning rate scheduler with predefined drops during training. Continuous training in the long context used the final learning rate from the 8K context, maintaining this rate throughout both training stages.

To evaluate the adaptation of the model, we used English PassKey\footnote{\href{https://github.com/CStanKonrad/long\_llama/blob/main/examples/passkey.py}{passkey.py}} and LongBench (v1)~\cite{bai2023longbench}. The LongBench evaluation set the maximum sample length according to the target context length, while the PassKey evaluation ranged from 8,000 to 128,000 tokens. Overall, the extension involved about 1.8 trillion tokens and tens of thousands of steps, but evaluations showed that it could be accomplished in just a few thousand steps.

\subsubsection{Post-train}

Each model trained on various versions of the post-train data (see the Section~\ref{subsec:post_train_data}) has its own hyperparameters, so in addition to several checkpoints within a single training (the model state is saved twice per epoch), we run several training iterations to select the best model from all of them. The final hyperparameters for the best open models are presented in Table~\ref{tab:sft_hyperparams}.

It is important to note that the final checkpoint does not always yield the highest performance metrics. In some versions of the dataset, the optimal model is achieved during the middle of the training process, while in others, it may be reached closer to the end. Therefore, selecting the best model involves a variety of heuristics based on specific needs. We choose from the metrics described in Section~\ref{subsec:benches}.

\begin{table*}[ht]
\small
\begin{tabular}{
>{}l 
>{}l 
>{}l 
>{}l 
>{}l}
{\textbf{model}}  & { \textbf{optimizer}} &  { \textbf{scheduler}} & { \textbf{params of scheduler}}  & { \textbf{hyperparameters}} \\
\toprule
{\gigainstruct} & {AdamW} & {Constant}  &{custom drop} & {-} \\ 
{\gigainstructdpo} & {AdamW} & {Cosine}  &{warmup: 200 steps, max steps: 7900} & {betas (0.9, 0.95), eps: 1.0e-8} \\ 

\bottomrule
\end{tabular}
\caption{Hyperparameters of the post-training models during the training.}
\label{tab:sft_hyperparams}
\end{table*}

\subsubsection{DPO}
\label{subsec:DPO}

In developing the \gigainstructdpo, we identified key issues with DPO, such as its focus on widening the gap between good and bad responses rather than improving accuracy, leading to hallucinations and instability. It also overlooks the importance of common token prefixes. To tackle these issues, we proposed modifications to the DPO loss function  (Equation~\ref{eq:dpo}), including unique weighting factors that prioritize enhancing good responses over suppressing bad ones, particularly concerning shared prefixes. We also added a normalized negative log-likelihood term relative to a reference model to stabilize loss ratios.

\begin{equation}
    \label{eq:dpo}
    \begin{split}
\text{loss} =\, &\mathbb{E}_{(x, y_w, y_l) \sim \mathcal{D}} \Bigl[ - \log \sigma \Bigl( \beta_w \log \frac{\pi_\theta(y_w \mid x)}{\pi_{\text{ref}}(y_w \mid x)} \\
&\quad - \beta_l \log \frac{\pi_\theta(y_l \mid x)}{\pi_{\text{ref}}(y_l \mid x)} \Bigr) + \log \frac{\pi_\theta(y_w \mid x)}{\pi_{\text{ref}}(y_w \mid x)} \Bigr]
    \end{split}
\end{equation}

\subsubsection{Optimal Tokenization}
\label{subsec:tokenization}
A new tokenizer has been developed to enhance the text encoding for Cyrillic words, programming languages, and LaTeX. We improve accuracy in handling code data by including common keywords and supporting spaces, tabs, and line breaks. High-frequency terms from LaTeX and programming are incorporated to minimize fragmentation, ensuring efficient tokenization of essential syntax elements. The selection of tokenizers was optimized to maximize the average length of tokens within domain-specific datasets.

\paragraph{Training Process}
We employed an iterative refinement process on a training dataset to maximize tokenization efficiency. Our focus was to ensure balanced performance across multiple domains, including programming languages such as C, Java, C\#, LaTeX markup, and general language corpora. The primary language of concern was Russian, with additional support for English and European languages, Arabic, Uzbek, and Kazakh. This effort primarily aims at the Russian community and the support of rarer languages, for which high-quality language models are scarce.

For training, we leveraged the Hugging Face Byte-Pair Encoding (BBPE) algorithm, conducting multiple experiments to generate candidate tokenizers. During these experiments, we gradually adjusted the proportion of texts from different domains (Russian, English, other languages, and code). This process resulted in a large number of candidate tokenizers (more than a hundred). From these, we selected the tokenizer that demonstrated the best performance compared to other tokenizers. The tokenizer training data and tokenizer comparison details are presented in Appendix \ref{appendix:tokenization}.

\section{Data}
\label{sub:data}

\subsection{Pre-train data}
\label{subsec:pre_train_data}

We aggregate diverse textual sources to construct a robust pre-training dataset, ensuring a balance between linguistic richness, domain-specific knowledge, and data quality. The dataset comprises 1) web-scraped texts, 2) high-quality publications, 3) programming code, and 4) synthetic data. The data statistic is presented in Table \ref{tab:pretraining_data}. We implement precise deduplication across all languages and sources to ensure corpus integrity and reduce redundancy. Additionally, we enhance the dataset for English-language data through MinHash deduplication~\cite{minhash}, which effectively minimizes semantic duplicates.

\paragraph{Web data}

To construct a high-quality pre-training corpus, we leverage Common Crawl web dumps from 2017-2023~\cite{frw}, \cite{dclm} and used a lightweight classifier~\cite{fasttext} to extract multilingual texts in Russian, English, Kazakh, Uzbek, Portuguese, and Arabic. These texts were further classified using LLMs and specialized models to identify educational~\footnote{\url{https://huggingface.co/datasets/HuggingFaceFW/fineweb-edu}} and high-value informational content~\cite{dclm}, resulting in 4.4T tokens of curated data. The dataset is predominantly English (63.76\%) and Russian (26.49\%), with Portuguese (7.80\%) and Arabic (1.90\%), and less than 0.06\% combined for Kazakh and Uzbek.

\paragraph{High-Quality Textual Sources}
We incorporate high-quality textual content from open-access books and academic articles, processed using advanced optical character recognition for accurate extraction. This adds 630B tokens of linguistic data. Additionally, we enrich the dataset with scientific and encyclopedic sources like arXiv, Wikipedia, and PubMed~\footnote{\url{https://pubmed.ncbi.nlm.nih.gov/download/}}, improving reasoning and factual consistency in the pre-training model.

\paragraph{Programming Code Corpus}
We use the StarCoder2~\cite{starcoder2} dataset alongside a curated set of open-source software code to create a diverse programming dataset that complies with licensing requirements. Machine learning models filter out low-quality code, yielding a 230B token subset ideal for code generation and understanding tasks.

\paragraph{Synthetic data}
Real-world data is limited by bias, privacy, and scarcity, while synthetic data is scalable and controlled. Phi-4~\cite{phi4} demonstrates that synthetic data pre-training improves performance on reasoning and STEM benchmarks. For math and programming, we built a Numina-inspired pipeline~\cite{numina} that expands seed mathematical problems by solving them multiple times and filtering via majority vote and threshold. We also created high-quality synthetic code tasks (complex Python problems with documentation, explanations, and assertions) with structured prompts and diversified them using  personas~\cite{personas} and lipograms~\footnote{\href{https://en.wikipedia.org/wiki/Lipogram}{https://en.wikipedia.org/wiki/Lipogram}}.

\begin{table}[t!]
    \small
    \centering
    \setlength{\tabcolsep}{2pt}
    \begin{tabular}{lcc}
    \toprule
       \textbf{Data source} & \textbf{Unique Tokens} & \textbf{Seen Tokens} \\
       \midrule
       Web & 4.4T & 5.6T \\
       HQ Sources & 630B & 1.3T \\
       Code & 230B & 1.3T \\
       Synthetic data & 9B & 81B \\
    \bottomrule
    \end{tabular}
    \caption{Pre-train data distribution.}
    \label{tab:pretraining_data}
\end{table}

\subsection{Post-train data}
\label{subsec:post_train_data}

Clean training data is essential during the post-training phase. All supervised fine-tuning dialogues are annotated by professional AI trainers who evaluate responses based on criteria like adherence to instructions, context awareness, factual accuracy, and safety. We created the Dialog Creation annotation project on the crowdsourcing platform Tagme~\footnote{\url{https://tagme.sberdevices.ru/}} to generate diverse dialogs across various domains while maintaining high data quality standards. AI trainers select the best responses from different model variants, using metadata for dataset balancing and error analysis to enhance model performance. To overcome the challenge of models retaining information from rare documents, we improved our model's memory and retrieval abilities through Retrieval-Augmented Generation following the experiments of the \citet{grattafiori2024llama3herdmodels}. This approach generates domain-specific training data from the pre-training corpus, enhancing contextual understanding.

Thus, the post-training of the open \gigainstruct model comprises about 250k items in the following proportion of data sources described in Table~\ref{tab:sftdata}.

\begin{table}[ht!]
\scriptsize
\centering
\setlength{\tabcolsep}{3pt}
\begin{tabular}{ll}
\toprule
\textbf{Domain} & \textbf{Proportion} \\
\midrule
chats & 10\% \\
long context (books) & 4\% \\
code & 4\% \\
science & 16\% \\
general world knowledge (web) & 34\% \\
translations & 1\%  \\
text editing & 12\% \\
business specifics & 3\% \\
functions / api & 16\% \\
\bottomrule
\end{tabular}
\caption{Post-training proportion of the task domains and instructions in the \gigainstruct.}
\label{tab:sftdata}
\end{table}

\section{Evaluation}
\label{sec:evaluation}

\begin{table*}[ht!]
\scriptsize
\centering
\begin{tabular}{l|l|lllll|ll}

\textbf{Benchmark} & \textbf{Shots} & \thead{GigaChat-\\A3B-instruct}  & \thead{GigaChat-A3B\\-instruct 1.5}  & \textbf{Qwen 2.5} & \textbf{T Lite} & \textbf{Llama 3.1} & \textbf{GigaChat2 Pro} & \textbf{GigaChat2 MAX} \\
\toprule
GSM8K &  5 & 0.764 & 0.774 & \underline{0.895} & 0.882  & 0.789 & 0.95 & \textbf{0.956}\\
MATH & 4 & 0.462 & 0.393  & \underline{0.704} & 0.592  & 0.329 & 0.752 & \textbf{0.773}\\
\midrule
HumanEval & 0 & 0.329 & 0.378 & \underline{0.854} & 0.799 & 0.683 & \textbf{0.915} & 0.871\\
MBPP & 0 & 0.385 & 0.441 & \underline{0.820} & 0.759 & 0.725 & 0.862 & \textbf{0.894} \\
\midrule
MMLU EN & 5 & 0.648 & 0.650 & \underline{0.710} & 0.718 & 0.682 & 0.821 & \textbf{0.86} \\
MMLU RU & 5 & 0.598 & 0.600 & \underline{0.632} & 0.626 & 0.569  & 0.775 & \textbf{0.805} \\
MMLU PRO EN & 5 & 0.348 & 0.357 &  \underline{0.565} & 0.509 & 0.443 & 0.644 & \textbf{0.667} \\
RUBQ & 0 & 0.675 & \underline{0.688} & 0.373 & 0.583 & 0.484 & 0.658 & \textbf{0.723 }\\
WINOGRANDE & 4 & 0.750 &\underline{0.762} & 0.636 & 0.670 & 0.624 & 0.796 & \textbf{0.832} \\
CyberMetric & 0 & 0.798 & 0.791 & 0.787 & \underline{0.883} & 0.796 & \textbf{0.84} & 0.832 \\
\midrule
IFEval & 0 & 0.411 & 0.433 & \underline{0.819} & 0.730 & 0.812 & 0.837 &  \textbf{0.899} \\
\bottomrule
\end{tabular}
\caption{Comprehensive comparison of models across Russian/English benchmarks. The best result in each column is highlighted in bold, the  best result in the same model size is underscored.}
\label{tab:benchmarks}
\end{table*}

\begin{table*}
\label{tab:mera_results}
\centering
\scriptsize

\begin{tabular}{l|c|ccccccc}  
\multirow{3}*{\textbf{Model}} & \multirow{3}*{\textbf{Total}} & \multirow{3}*{\textbf{RWSD}} & \multirow{3}*{\textbf{ruModAr}} & \multirow{3}*{\textbf{USE}} & \multirow{3}*{\textbf{MaMuRAMu}} & \multicolumn{3}{c}{\textbf{ruHHH}} \\ 
\cmidrule(lr){7-9}
 & & & & & & \textbf{Honest} & \textbf{Helpful} & \textbf{Harmless} \\
\toprule
Human Benchmark & 0.852 & 0.835 & 0.942 & 0.701 & 0.796 & 0.705 & 0.797 & 0.948 \\
\midrule
Claude 3.7 Sonnet & \textbf{0.682} & \textbf{0.788} & 0.919 & 0.536 & \textbf{0.89} & 0.82 & \textbf{0.864} & 0.931 \\
\textbf{GigaChat 2 MAX} & 0.67 & 0.642 & \textbf{0.963} & \textbf{0.581} & 0.864 & 0.803 & 0.831 & \textbf{0.948} \\
Gemini 1.5 Pro & 0.675 & 0.627 & 0.707 & 0.433 & 0.868 & 0.836 & 0.797 & 0.931 \\
GPT-4o & 0.642 & 0.496 & 0.729 & 0.457 & 0.874 & \textbf{0.852} & 0.729 & 0.862 \\
DeepSeek V3 & 0.677 & 0.612 & 0.718 & 0.499 & 0.882 & 0.803 & 0.763 & 0.793 \\
\midrule
Phi-3.5-MoE-Inst & 0.487 & 0.465 & 0.464 & 0.199 & 0.726 & 0.656 & 0.644 & 0.81 \\
\textbf{GigaChat 2 Pro} & \textbf{0.649} & 0.665 & \textbf{0.943} & \textbf{0.534} & 0.831 & 0.803 & 0.814 & 0.897 \\
Mixtral-8x22B-Inst & 0.486 & 0.473 & 0.523 & 0.269 & 0.747 & 0.836 & \textbf{0.881} & \textbf{0.966} \\
Qwen2.5-72B-Inst & 0.601 & \textbf{0.715} & 0.665 & 0.32 & 0.849 & \textbf{0.869} & 0.831 & 0.897 \\
Llama-3.1-405B-Inst & 0.59 & 0.677 & 0.573 & 0.357 & \textbf{0.868} & 0.803 & 0.864 & 0.759 \\
\midrule
RuadaptQwen2.5-7B & 0.536 & 0.465 & 0.492 & 0.162 & 0.751 & 0.738 & 0.78 & 0.776 \\
\textbf{GigaChat 2} & 0.541 & 0.369 & \textbf{0.854} & \textbf{0.361} & 0.766 & 0.754 & 0.814 & \textbf{0.931} \\
T-lite-it-1.0 & 0.552 & 0.535 & 0.493 & 0.147 & 0.775 & 0.689 & 0.797 & 0.862 \\
\textbf{\gigainstruct} & 0.512 & 0.535 & 0.853 & 0.325 & 0.728 & 0.689 & 0.78 & 0.759 \\
\textbf{\gigainstruct 1.5} & 0.511 & 0.512 & 0.84 & 0.32 & 0.728 & 0.689 & 0.831 & 0.793 \\
gemma-3-27b & \textbf{0.567} & \textbf{0.588} & 0.626 & 0.328 & \textbf{0.797} & \textbf{0.82} & \textbf{0.864} & 0.914 \\
\midrule
gemma-2-9b & \textbf{0.453} & 0.558 & 0.592 & \textbf{0.154} & \textbf{0.689} & 0.574 & 0.627 & 0.552 \\
\textbf{\gigabase} & 0.422 & 0.508 & \textbf{0.608} & 0.127 & 0.675 & 0.574 & 0.593 & 0.552 \\
Llama-3.2-3B & 0.362 & 0.477 & 0.592 & 0.075 & 0.528 & 0.41 & 0.542 & 0.483 \\
Yi-1.5-9B-32K & 0.428 & 0.569 & 0.516 & 0.12 & 0.516 & \textbf{0.59} & \textbf{0.661} & \textbf{0.621} \\
Qwen1.5-7B & 0.374 & 0.558 & 0.485 & 0.056 & 0.52 & 0.541 & 0.627 & 0.603 \\
Mistral-7B-v0.1 & 0.404 & \textbf{0.581} & 0.517 & 0.107 & 0.585 & 0.574 & 0.559 & 0.552 \\
ruGPT-3.5 & 0.213 & 0.462 & 0.001 & 0.082 & 0.226 & 0.459 & 0.475 & 0.483 \\
\bottomrule
\end{tabular}
\caption{MERA benchmark results. The model's descriptions are available in the \href{https://mera.a-ai.ru/en/text/leaderboard}{MERA leaderboard}}
\end{table*}

\subsection{Benchmarks}
\label{subsec:benches}

For the evaluation of the models, we use various common benchmarks in English and Russian that assess skills such as Mathematics Performance (GSM8K~\cite{cobbe2021training}, MATH~\cite{hendrycks2021measuring}), Coding Ability (HumanEval~\cite{chen2021evaluating}, MBPP~\footnote{\url{https://github.com/google-research/google-research/tree/master/mbpp}}), General Knowledge (MMLU EN~\cite{hendrycks2020measuring}, MMLU RU~\footnote{\url{https://mera.a-ai.ru/ru/tasks/9}}, MMLU PRO~\cite{wang2024mmlu}, RUBQ~\cite{korablinov2020rubq}, WINOGRANDE~\cite{sakaguchi2021winogrande}), Cybersecurity Knowledge (CyberMetric~\cite{tihanyi2024cybermetric}), and Instruction Following (IFEval~\cite{zhou2023instruction}). Table~\ref{tab:benchmarks} presents a comprehensive performance comparison between open versions of GigaChat models and other open post-trained LLMs of compatible sizes (Llama 3.1 8b~\footnote{\url{https://huggingface.co/meta-llama/Llama-3.1-8B}}, Qwen 2.5 7~\footnote{\url{https://huggingface.co/Qwen/Qwen2.5-7B-Instruct}}, and T-Lite~\footnote{\url{https://huggingface.co/AnatoliiPotapov/T-lite-instruct-0.1}}) across benchmarks. 
As the benchmark was created specifically for the Russian language, we present the assessment of pre-training and instructing models on the benchmark MERA~\cite{fenogenova2024mera}. For all tests, the LM Evaluation Harness framework \footnote{\url{https://github.com/EleutherAI/lm-evaluation-harness}} was used. 

\subsection{Results}
\label{subsec:results}
\paragraph{English Benchmarks}
The \gigainstruct and \gigainstructdpo models (3.3B active parameters) show a balanced trade-off between scale and performance against larger 7–8B counterparts (Qwen2.5 7B, Llama 3.1 8B, T-Lite). While mathematical (-14\% GSM8K, -34\% MATH) and programming (-46\% MBPP, -55\% HumanEval) gaps reflect parameter limitations, they excel in reasoning (+15\% RUBQ, +12\% WINOGRANDE) and retain competitiveness in MMLU (-5\% to -8\%). Challenges in high-difficulty MMLU PRO (-36\%) and instruction following (-47\% IFeval) persist, though DPO optimization yields targeted improvements. For the CyberMetric benchmark, new models also show competitive results, being 11\% lower than the leader. Concerning \gigamaxnew, \gigapronew, the models show the best scores for all benchmarks, slightly falling short only on CyberMetric (-5\%).

\paragraph{Russian Benchmarks}
Designed for Russian-language proficiency, the models (\gigamaxnew, \gigapronew, \giganew) achieve near-state-of-the-art results on MERA benchmark (±2–7\%) and dominate specialized tasks: ruModAr (+4\% to +29\%) and USE (+7\% to +33\%) highlight strengths in logic and complex comprehension. Coreference resolution (RWSD: -7\% to -18\%) and advanced reasoning (MaMuRAMu: -3\% to -4\%) show room for growth, yet performance remains competitive against both frontier models (e.g., GPT-4) and mid-tier alternatives. \gigainstruct and \gigainstructdpo show a performance close to GigaChat 2. \gigabase reaches the level of best 9 billion pre-train models trailing by 12-20\% on RWSD and USE, by 2\% on MaMuRAMu, leading by 2\% on ruModAr. Concerning the ruHHH dataset aimed at scoring the model's ability to determine the Honest, Helpful and Harmless behavior all GigaChat models show nearly the highest results among the same tier models: \gigamaxnew, \gigapronew, \giganew show the best or nearly the best scores for Harmless while being slightly behind the leaders for Honest and Helpful (-9\% to -4\%); \gigabase remains competitive against the other 3B--13B models (-12\% to -3\%); \gigainstruct, \gigainstructdpo show close scores while demonstrating that DPO may help determine Helpful behavior better (+7\% compared to without DPO).

\section{Conclusion}
We present the GigaChat family of LLMs, which is the only model developed from scratch during the pre-training stage specifically for the Russian language. By employing the MoE architecture and a specialized tokenizer, we have developed models that effectively address Russian linguistic and cultural nuances while achieving competitive performance against leading benchmarks. Our open-source release of three GigaChat models and user-friendly interfaces like a Telegram bot and a Web application for the frontier models aims to encourage further research and industrial applications in Russian NLP. The contributions outlined, including the introduction of Russian-focused models and experimental results, reflect our commitment to enhancing the field. By providing these resources to the community, we hope to foster innovation and collaboration in developing inclusive and effective language technologies for Russian-speaking users.

\section*{Ethical Statement}
\label{sec:ehtical}

\paragraph{Possible Misuse} Our research should not contribute to creating content that negatively impacts individual or community well-being. This includes the following restrictions: (i) involvement in legislative applications or censorship, (ii) dissemination of disinformation or infringement on the right to access information, (iii) dehumanizing or misrepresenting individuals or their religions, cultures, or beliefs, and (iv) promoting harmful or discriminatory content. To address this issue, the models' API format includes a censorship filter to mitigate inappropriate content that could pose potential risks.

\paragraph{Biases and data quality} The pre-training data for all the models includes a wide range of content from Russian and English internet sources, which may introduce various stereotypes and biases. Thorough evaluations of these models are crucial to identifying potential vulnerabilities when applied to data outside their training domain.

\paragraph{Energy Efficiency and Usage} 
We compute the $CO_2$ emissions from training our LLMs as Equation~\ref{eq:co2}~\cite{strubelletal2019energy}:

\vspace{-2pt}
\begin{equation}
\label{eq:co2}
    CO_2 = \frac{PUE * kWh * I^{CO2}}{1000}
\end{equation}

The resulting number of the $CO_2$ for the open models is presented in Table~\ref{tab:co2}. 251k kg of $CO_2$ is approximately equivalent to a round-trip flight from New York to London emits ~1,600 kg of $CO_2$ per passenger.

\begin{table}[t!]
    \small
    \centering
    \setlength{\tabcolsep}{2pt}
    \begin{tabular}{lc}
    \toprule
       \textbf{Model} & \textbf{$CO_2$ (kg)} \\
       \gigabase & 251k \\
       \gigainstruct & 253k \\
       \gigainstructdpo & 255k \\
    \bottomrule
    
    \end{tabular}
    \caption{$CO_2$ emissions of the models training.}
    \label{tab:co2}
\end{table}

\section*{Limitations}
\label{sec:limitations}


\paragraph{Lack of Reasoning Capabilities} The models do not exhibit advanced reasoning abilities (like the models like DeepSeeek R1), which may restrict its effectiveness in tasks requiring complex problem-solving or logical inference.

\paragraph{Alignment Preferences} The models have been specifically aligned to generate long and aesthetically pleasing chat responses. While this may appeal to some users, others might find such responses verbose or less practical for their needs. 

\paragraph{Tokenizator} The effectiveness of the trained tokenizer and the trained LMs is highly dependent on the quality and size of the corpus used. A limited or biased corpus can lead to suboptimal tokenization and model performance,  potentially missing critical linguistic nuances and specific domain cases, such as characters from formal or other languages.

\paragraph{Reproducibility Issues} Due to the use of closed pre-training, fine-tuning, and DPO datasets for proprietary models, the results cannot be independently replicated or verified. This lack of transparency may inhibit further research and validation efforts. However, we are open-sourcing three versions of the MoE-based GigaChat, and we hope this will encourage further research in Russian.

\section*{Acknowledgments}
We extend our deepest gratitude to all members of the GigaChat Team for their tireless efforts and dedication to this ambitious project. Their collective expertise and collaboration were instrumental in bringing this research to life. 
We are especially thankful to Sergey Markov and the R\&D team for their invaluable feedback and contributions during the early stages of this work. Our sincere appreciation also goes to Andrey Belevtsev for his visionary insights and strategic guidance, which helped shape the direction of this project. We sincerely acknowledge Anton Chigin and Sergei Galustian for their contributions to runtime and architecture development.
Additionally, we acknowledge the many technical teams whose infrastructure and engineering excellence enabled the development, hosting, and deployment of the system for users.

\paragraph{Author Contributions}
\begin{itemize}[nosep]
    \item \textit{Administration and Supervision}: Fedor Minkin
    \item \textit{Pre-training Team. Data}: Ivan Baskov, Valeriy Berezovskiy, Dmitry Kozlov, Ainur Israfilova, Lukyanenko Ivan
    \item \textit{Pre-training Team. Training}: Gregory Leleytner, Evgenii Kosarev, Mamedov Valentin
    \item \textit{Supervised Fine-Tuning (SFT) Team. Training}: Gregory Leleytner, Emil Shakirov, Smirnov Daniil, Mikhail Kolesov, Kolodin Egor, Aleksandr Proshunin
    \item \textit{Supervised Fine-Tuning (SFT) Team. Data}: Nikita Savushkin, Eldar Damirov, Daria Khomich, Daria Latortseva, Sergei Porkhun,  Yury Fedorov, Oleg Kutuzov, Polina Kudriavtseva, Sofiia Soldatova, Stanislav Pyatkin, Dzmitry Menshykh, Grafov Sergei, Karlov Vladimir, Ruslan Gaitukiev, Arkadiy Shatenov
    \item \textit{Evaluation and Metrics}: Emil Shakirov, Smirnov Daniil, Artem Orlov, Alena Fenogenova
    \item \textit{Tokenization}: Sergei Averkiev
    \item \textit{Expert Interpretations}: Ilya Shchuckin
    \item \textit{Research Supervision and Coordination}: Alena Fenogenova
\end{itemize}

\bibliography{custom}

\appendix

\section{Appendix}
\label{sec:appendix}

\subsection{Training details}
\label{subsec:details}

We used a mixed precision training methodology (bfloat16 for most operations and fp32 for critical components, such as the router). The complete training process accumulated approximately 10 trillion tokens, with the final annealing phase comprising 40 billion tokens of pre-trained data described in Section~\ref{subsec:pre_train_data}.

We tackled communication bottlenecks in large-scale distributed training environments with over 256 GPUs by increasing batch size instead of adding more devices with the same workload. This strategy allowed for overlapping communication and computation, minimizing idle time and enhancing training throughput. The sparse computation patterns of the MoE architecture, along with a moderate hidden size, enabled us to significantly increase the batch size per device while staying within memory limits.

Throughout the training process, we systematically monitored expert utilization and router confidence using entropy-based metrics: $H\_utilization$ (quantifying token distribution between experts) and $H\_sparsity$ (measuring router confidence). We analyzed token distribution among experts and monitored $top-k$ router scores, identifying several critical issues: \textit{expert collapse phenomena} (experts receiving minimal token assignments), \textit{disproportionate token} processing by specific experts, and \textit{router uncertainty} indicated by consistently low confidence scores. These metrics guided our hyperparameter optimization, especially for the auxiliary load balancing loss for uniform expert utilization. Visualizing expert utilization patterns offered insights that shaped our decision to implement a standard Gated MLP in the first layer.

\subsection{Ablation study: Expert interpretations}
\label{subsec:experts}

During the experiments on the model architecture, we analyze router behavior to investigate if experts in \gigabase, specialize in specific domains such as math, medicine, and code. To do this, we constructed embeddings for a subset of the Pile~\cite{pile} dataset~\footnote {We use the version~\url{https://huggingface.co/datasets/monology/pile-uncopyrighted} of the set where all copyrighted content was removed} using router activations. Each embedding $emb$ is a matrix of size $l \times e$, where $l$ is the number of MoE layers and $e$ is the number of experts in one layer (not including shared experts). Each sample $emb_{ij}$ is calculated as the number of activations of expert $j$ in layer $i$ normalized by the length of the sample in tokens.

\begin{figure*}[ht]
   \centering
  \includegraphics[width=\textwidth]{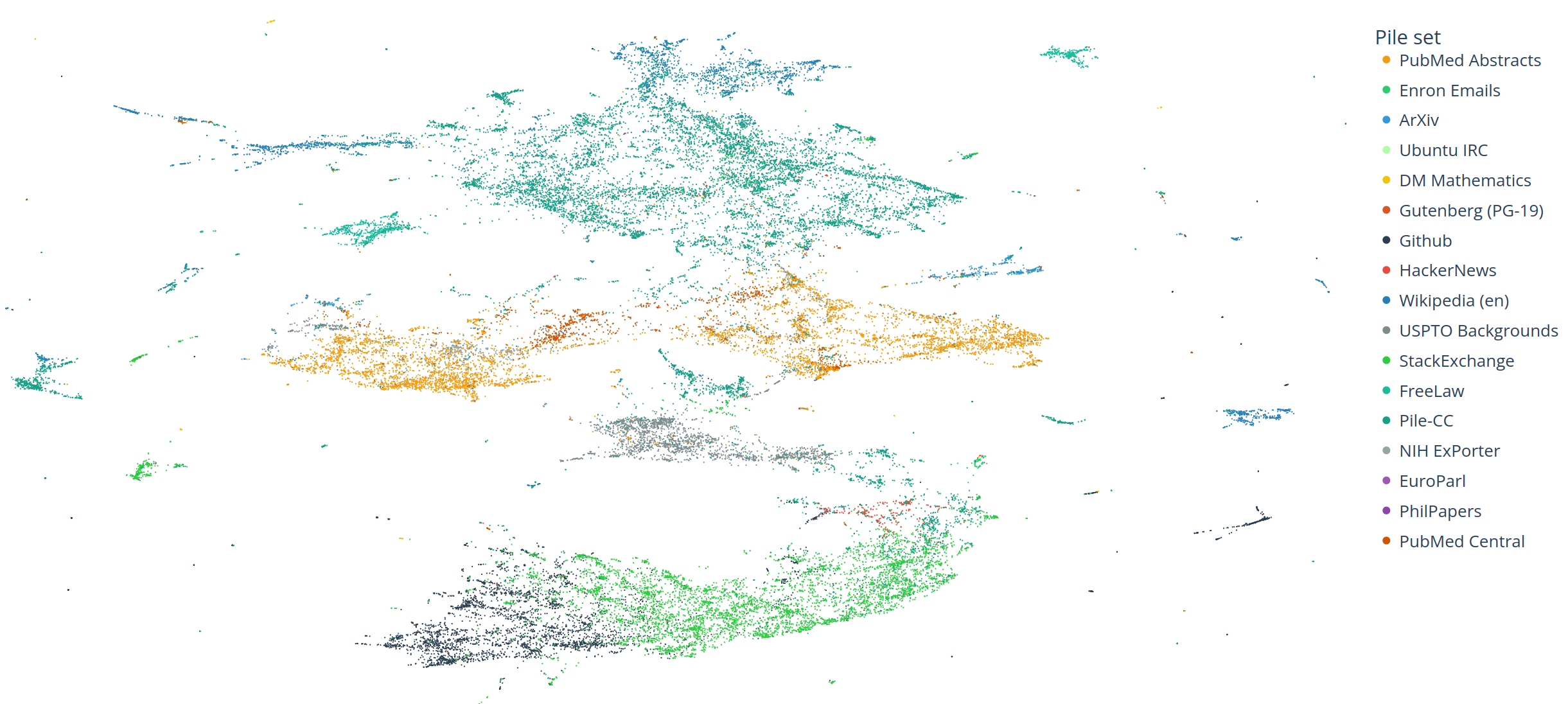}
  \caption{2d-projection of embeddings with UMAP}
  \label{fig:clusters}
\end{figure*}

We clustered the embeddings with UMAP and HDBSCAN, revealing that samples grouped by domain (Fig.~\ref{fig:clusters}), indicating that router decisions encode domain information. This aligns with the findings in~\cite{li2024mixtureofexpertsllmsecretlyembedding}, where MoE models provided effective embeddings without the need for fine-tuning. Clusters were identified in sports, cooking, biology, and programming domains.
 
We created domain-specific embeddings by averaging values within clusters. These embeddings help differentiate experts in those fields. To identify significant experts, we set values below \(\frac{3}{e}\) to zero, keeping only those at least three times greater than expected. We then use filtered embeddings to guide our model toward specific domains by adjusting router activations to prioritize selected experts.

We found that this method allows us to control generation flow~\footnote{Examples of the code for generation control are presented in the \href{https://github.com/Elluran/concentration\_notebooks/blob/main/steering.ipynb}{example notebook}.}; for example, using sports-related embeddings led to texts focused on sports. Similar patterns emerged in other domains. While this method has potential benefits, it also has limitations that may hinder the model's language modeling capabilities. Despite these challenges, we view this approach as promising and intend to provide a more detailed analysis in future research.

\subsection{Tokenizer details}
\label{appendix:tokenization}

For tokenizer training, we utilized both open-source datasets, namely FRW~\cite{refinedweb}, RedPajama~\cite{together2023redpajama}, StarCoder~\cite{li2023starcoder}, as well as collected from the Web like Common Crawl~\footnote{\href{https://commoncrawl.org/get-started}{https://commoncrawl.org/get-started}}, Wikipedia~\footnote{\href{https://dumps.wikimedia.org/ruwiki/latest/}{https://dumps.wikimedia.org/ruwiki/latest/}} and Stack Exchange~\footnote{\href{https://archive.org/details/stackexchange}{https://archive.org/details/stackexchange}}. For details on post-processing and cleaning the open-source datasets, refer to their respective articles. We filtered the datasets using established heuristics, such as language-based filtering and removing personal information, promotional content, and duplicates. Several sets of data were prepared for training tokenizers, varying in size from 30 billion to 300 billion characters to reflect different text lengths.

To ensure the effectiveness of our approach, we tested tokenizers against established models, including GPT-4, GPT-4o, Mistral, Qwen2, and DeepSeek. The comparison was based on the average character-per-token ratio across different domains, as summarized in Table~\ref{tab:tokenizers} with selected domains. Tokenizers with the prefix giga\_tokenizer represent multiple variants from our experiments, differing in data balancing strategies and the number of additional tokens introduced.

\begin{table*}[ht]
    \centering
    \begin{tabular}{ccccccccc}
        \toprule
        \multirow{3}*{\textbf{Tokenizer}} & \multicolumn{3}{c}{\textbf{Languages}} & \multirow{3}*{\textbf{ArXiv}} & \multicolumn{3}{c}{\textbf{Wiki}} & \multirow{3}*{\textbf{Mean Score}} \\
        \cmidrule(lr){2-4} \cmidrule(lr){6-8}
          & C & Java & C\# &  & Ru & Ar & En &  \\
        \midrule
        giga\_tokenizer\_1  & 3.57 & 4.15 & 4.62 & \textbf{3.61} & \underline{4.18} & \underline{3.34} & 4.47 & \textbf{3.99} \\
        giga\_tokenizer\_2  & 3.56 & 4.14 & 4.60 & \textbf{3.61} & 4.14 & 3.30 & 4.44 & \underline{3.97} \\
        gpt-4o             & \underline{3.74} & 4.43 & 4.88 & 3.39 & 3.40 & 3.07 & \textbf{4.68} & 3.94 \\
        giga\_tokenizer\_5  & 3.39 & 3.97 & 4.44 & \underline{3.54} & \textbf{4.20} & \textbf{3.50} & 4.43 & 3.92 \\
        giga\_tokenizer\_3  & 3.51 & 4.11 & 4.59 & \underline{3.54} & 4.04 & 3.25 & 4.35 & 3.91 \\
        giga\_tokenizer\_4  & 3.50 & 4.11 & 4.58 & 3.53 & 4.00 & 3.21 & 4.33 & 3.90 \\
        llama-3            & \textbf{3.75} & \underline{4.54} & \textbf{4.99} & 3.38 & 3.02 & 2.60 & 4.62 & 3.85 \\
        mistral-nemo       & 3.38 & 4.06 & 4.50 & 3.49 & 3.18 & 3.24 & 4.51 & 3.76 \\
        qwen2              & 3.69 & 4.52 & 4.95 & 3.31 & 2.70 & 2.56 & 4.50 & 3.75 \\
        gpt-4              & \underline{3.74} & \textbf{4.55} & \underline{4.98} & 3.38 & 2.04 & 1.44 & \underline{4.62} & 3.54 \\
        nemotron-4-256k    & 2.82 & 3.34 & 3.76 & 3.25 & 3.20 & 2.93 & 4.57 & 3.41 \\
        deepseek-coder-v2  & 2.95 & 3.51 & 3.92 & 3.35 & 2.39 & 1.11 & 4.42 & 3.10 \\
        deepseek-v2        & 2.95 & 3.51 & 3.92 & 3.35 & 2.39 & 1.11 & 4.42 & 3.10 \\
        mistral-large      & 2.75 & 3.26 & 3.64 & 3.14 & 2.46 & 1.13 & 4.04 & 2.92 \\
        \bottomrule
    \end{tabular}
    \caption{Comparison of Tokenizers by Character-per-Token Ratio. }
    \label{tab:tokenizers}
\end{table*}

\end{document}